\documentclass[runningheads]{llncs}
\usepackage{graphicx}
\usepackage{todonotes}
\usepackage{ulem}
\usepackage{array}
\usepackage{booktabs}
\usepackage{layouts}
\usepackage{ntheorem}
\theoremseparator{:}

\begin{document}
\title{Exploring connections of spectral analysis and transfer learning in medical imaging}
%
%
\ifdefined\DOUBLEBLIND
    \author{Anonymous Authors}
    \institute{Institutes}
\else
    \author{Yucheng Lu\inst{1} \and
    Dovile Juodelyte\inst{1} \and \\
    Jonathan D. Victor\inst{2} \and
    Veronika Cheplygina\inst{1}
    }
    \authorrunning{Y. Lu et al.}
    %
    \institute{IT University of Copenhagen, Copenhagen, Denmark, 2300\\ \email{\{yucl,doju,vech\}@itu.dk} \and
    Weill Cornell Medicine, New York, USA, 10065 \\ \email{jdvicto@med.cornell.edu}
    }
\fi

\maketitle              
\begin{abstract}
In this paper, we use spectral analysis to investigate transfer learning and study model sensitivity to frequency shortcuts in medical imaging. By analyzing the power spectrum density of both pre-trained and fine-tuned model gradients, as well as artificially generated frequency shortcuts, we observe notable differences in learning priorities between models pre-trained on natural vs medical images, which generally persist during fine-tuning. We find that when a model’s learning priority aligns with the power spectrum density of an artifact, it results in overfitting to that artifact. Based on these observations, we show that source data editing can alter the model's resistance to shortcut learning. 

\end{abstract}
\section{Introduction}
\label{sec:intro}
Deep learning has achieved many advances in medical image classification, even showing performances on par with medical experts. However, convolutional neural networks (CNNs) may be prone to shortcut learning \cite{geirhos2020shortcut}, such as surgical markers \cite{winkler2019association}. As a consequence, instead of capturing the semantic contents, the model makes predictions based on the shortcuts, which, in the worst case, leads to unreliable results if their association with semantics differs between the training dataset and the images used in real-world applications. 

Most studies investigate shortcut learning in the context of training from scratch. However, little is understood about the importance of shortcuts in transfer learning, which is crucial in the medical domain for two reasons. First, transfer learning is often involved in medical image analysis due to the limited amount of labeled data \cite{pan2010survey,cheplygina2019cats,raghu2019transfusion}. Second, next to obvious shortcuts like pen markings, CT and MR scans, in particular, can have subtle shortcuts in the spectrum domain that may not be noticed by the human eye. This prompts us to explore the sensitivity of transfer learning to spectral shortcuts in medical image classification tasks and how to mitigate the negative impacts it brings about. 

To this end, we use spectral analysis to investigate the role of power spectrum density (PSD) in pre-training and fine-tuning and observe distinct differences in their learning priorities, which are related to shortcut learning. Based on these observations we show through experiments that resistance to common detrimental frequency shortcuts could be altered via source data editing. In summary, our main contributions are as follows:

\begin{itemize}
    \item We apply spectral analysis to transfer learning. Specifically, we use learning priority to analyze a model's frequency bias before and after fine-tuning and reveal distinct differences between natural and medical images.

    \item We study the relationship between the pre-trained and the fine-tuned models in terms of PSD, and observe that the fine-tuned model robustness is related to the pre-trained model learning priority.

    \item We show that the PSD of a pre-trained model can be altered through source data editing, which could lead to greatly improved robustness against shortcut learning.
\end{itemize}

\section{Related Work}
\label{sec:relatedwork}
The standard transfer learning protocol involves fine-tuning a pre-trained model on a small target dataset, where the pre-trained weights are determined by training on a large-scale source dataset, with ImageNet \cite{deng2009imagenet} being a popular choice. However, due to the domain gap between ImageNet and target medical datasets, several studies showed that pre-training on medical data can improve performance \cite{xie2018pre,ghesu2201self,cheplygina2019cats}, leading to a large-scale radiological image dataset RadImageNet \cite{mei2022radimagenet}. Another factor related to model performance in transfer learning is shortcut learning. For example, \cite{juodelyte2024source} found that RadImageNet is more robust to shortcuts, such as those relying on noise and denoising. Since we notice that these shortcuts are \textit{spectrum-related}, and the spectra of natural and medical images have distinct differences \cite{xu2019systematic}, we classify related work into three categories: frequency bias, frequency shortcuts, and spectrum augmentation.

\subsubsection{Frequency bias}
After \cite{yin2019fourier} showed that CNNs often learn a stronger bias towards a frequency band that is highly correlated with the characteristics of image degradation, many researchers further investigated this phenomenon. \cite{wang2020high} found that while humans make use of semantic content, powerful CNNs tend to rely on high-frequency components, leading to lower robustness. Their experiments also revealed a trade-off between the model's robustness and performance. Similarly, \cite{chen2021amplitude} pointed out that this counter-intuitive behavior is related to an over-emphasis on amplitude, compared to relative phase. \cite{abello2021dissecting} examined the importance of each frequency band based on an energy distribution model to control the ratio between classification performance deterioration and image quality degradation for each band. They suggested that the frequency bias is not tightly related to CNN architectures or model depth but to discriminative features in the mid-frequency band. \cite{lin2022investigating} further observed that the preference for low-to-mid-frequency is due to the considerable suppression of high-frequency bands in feature extraction.

While these studies showed the presence of frequency bias, they only considered natural images. In comparison, we analyze frequency bias in both natural image and medical image domains and observe distinct differences.

\subsubsection{Frequency shortcut}
Previous work showed that CNNs are often biased towards specific bands, learning to recognize signature patterns in the spectrum, especially when these patterns are confounders.  \cite{wang2022frequency,wang2023neural} offered a deep understanding of these frequency shortcuts by selectively extracting spectral patterns while disregarding irrelevant frequencies. Their experiments showed that during initial training, CNNs seek to find simple solutions, such as the most distinct spectrum characteristics. These shortcuts are not limited to low-frequency bands but can also be high-frequency ones. Building on these findings, \cite{wang2023dfm} proposed a method to mitigate shortcut learning by filtering samples of each class based on the frequency content of the other classes.

Here, we focus on the model's robustness against frequency shortcuts in transfer learning. Particularly, we are interested in exploring the link between the kernels' frequency response and their initial weights as determined from different source datasets. Our study reveals that the fine-tuned model robustness relates to the pre-trained model's learning priority.

\subsubsection{Spectrum augmentation} It is difficult to improve the learning of domain-invariant features via data augmentation in the spatial domain alone \cite{yin2019fourier}. An alternative way to mitigate frequency bias is to apply data augmentation in the frequency domain. Several variations on this theme have been pursued. \cite{fridovich2022spectral} studied the frequency decomposition of learned functions by adding noise to various frequencies via label smoothing. They observed that a high-accuracy model responds well to high frequencies across classes but focuses more on low frequencies within each class. \cite{cheng2023frequency} exploited the impact of frequency components in few-shot learning, where a class-discriminative filtering scheme based on Grad-CAM \cite{selvaraju2017grad} was applied to the training samples to enhance the model's ability to capture task-relevant frequencies. \cite{xu2021fourier} blended the amplitude components from two images while keeping the phase components unaltered based on their semantic-preserving property; while \cite{ngnawe2023robustmix} fused the spectra of two random classes with independent focuses on low and high frequencies and trained the model to predict the weighted probabilities for the two classes, retaining a controllable sensitivity across various frequency bands. \cite{yucel2023hybridaugment++} swapped the frequency bands between two randomly selected images or augmented variants of a single image and further applied phase perturbation from a third image to enrich the data augmentation.

Together, these studies provided new approaches to data augmentation but did not answer the question of how transfer learning might benefit -- which we address here. Our experiments show that data editing in the source domain affects the model's robustness against shortcut learning in the target domain.

\section{Methodology}
\label{sec:method}
\subsubsection{Datasets and models}
As \textit{sources} we use ImageNet \cite{deng2009imagenet} and RadImageNet \cite{mei2022radimagenet}. ImageNet has 1.2M training and 50K validation images in 1K classes, while RadImageNet has 1M  training and 112K validation images in 165 classes. We pre-train a ResNet-50 \cite{he2016deep} (implementation details in Supplementary) as it is a common choice for medical images. As \textit{targets} we select two small medical datasets: LoDoPaB-CT \cite{leuschner2021lodopab} -- a subset of LIDC-IDRI \cite{armato2011lung}, and KneeMRI \cite{vstajduhar2017semi}. We chose these datasets as both of their imaging pipelines involve frequency-domain reconstruction. To simplify the analysis of frequency shortcuts, we binarize the tasks to benign (malignancy score $<$ 3) vs malignant for LoDoPaB-CT, and healthy vs injured ligament for KneeMRI. This results in the following train/validation/test partitions: 375/125/1033 samples (198/66/548 studies) for LoDoPaB-CT, and 375/125/871 samples (375/125/582 studies) for KneeMRI.

\subsubsection{Frequency shortcuts}
We introduce shortcuts by altering the images in two frequency-related ways: noise and denoising -- here denoted as ``artifacts''. The noise level in CT images varies because of automatic exposure control and the choice of reconstruction filters. Denoising is commonly applied after reconstruction as a spatial filtering operation, but the extent of denoising can vary from image to image. Thus, both result in alterations of the frequency content of the image and could lead to frequency shortcuts.

We select projection-domain Photon noise in CT \cite{leuschner2021lodopab} and non-local means (NLM) denoising in MRI \cite{manjon2008mri} because they have distinct spectral statistics. To create a spurious correlation between the artifacts and the labels, we add the artifacts to all negative samples in the test set and a certain amount (e.g. 50\%) of positive samples in the training set. This design ensures that if the model detects the shortcut, its out-of-distribution (O.O.D.) performance will decrease, while the independent-and-identically-distributed (I.I.D.) performance will improve.

\subsubsection{Power spectrum density}
\begin{figure}[t]
    \centering
    \includegraphics[scale=1]{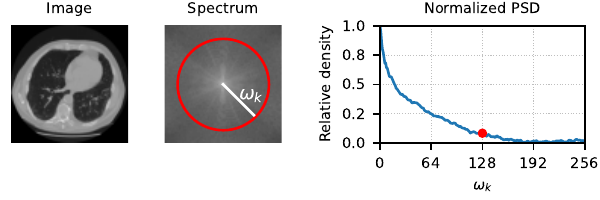}
    \caption{Example of a PSD. From left to right: original image, its spectrum with a selected frequency $\omega_{128}$, and the PSD with the highlighted frequency $\omega_{128}$.}
    \label{fig-01}
\end{figure}

To characterize the statistics of datasets and model weights in the frequency domain, we convert the standard 2D spatial power spectrum into a 1D PSD by integrating the spectrum values over all angles. The resulting quantity provides a comprehensive measure of power distribution across frequencies and is especially useful when an artifact or a feature lies in a specific frequency band. The PSD is computed as follows:

\begin{equation}
    PSD\left ( \omega_{k} \right ) = \int_{0}^{2\pi}\left \| \mathcal{F}\left ( X \right ) \left ( \omega_{k}\cos\phi,\omega_{k}\sin\phi \right ) \right \|d\phi,
    \label{eq-01}
\end{equation}
where $\omega_{k}$ represents radial frequency, $k \in \left \{ 0,1,\cdots,\frac{1}{2}M - 1 \right \}$, $M$ is the input size (assuming square shape). $\phi$ is the angle, and $\mathcal{F}$ is the Fourier transform. An example of PSD is presented in Fig.~\ref{fig-01}.

It is worth noting that PSD is versatile. When the input $X$ is image data, the PSD shows the overall spectral distribution. To analyze a trained model, one can compute the gradient map back-propagated from the prediction loss to the input image as $X$ to analyze the model's spectral \textit{learning priority} \cite{lin2022investigating}.

\section{Experimental results}
\label{sec:experiment}
\subsubsection{ImageNet is prone to shortcut learning}
\begin{figure}[t]
    \centering
    \includegraphics[scale=1]{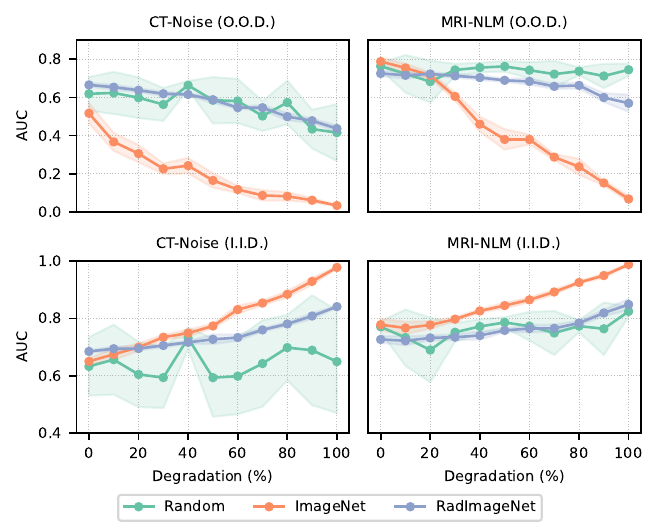}
    \caption{Baseline results (mean and standard deviation of AUC across 5-folds) as a function of degradation (amount of artifacts in the training set), performance on O.O.D. (top) and I.I.D. (bottom) test sets.}
    \label{fig-02}
\end{figure}

We pre-trained the model on the original ImageNet and RadImageNet and fine-tuned it on the target datasets as the baseline. The 5-fold cross-validation results are shown in Fig.~\ref{fig-02}. RadImageNet has higher robustness against frequency shortcuts, whereas ImageNet exhibits poor generalization ability when tested on O.O.D. images. In comparison, random initialization (i.e. training from scratch, dubbed ``random'') shows dramatic fluctuation in performance across folds, indicating its instability on small datasets. However, both ImageNet and RadImageNet pre-trained models have competitive performance on I.I.D. data, which reveals that the source dataset plays an important role in shortcut learning.

\subsubsection{Learning priority is stable during transfer}
We computed the PSDs of models (i.e. learning priorities) pre-trained on ImageNet and RadImageNet. The results are plotted in Fig.~\ref{fig-03} (top row). We notice that ImageNet pre-trained model has higher gradients in the mid-to-high frequency bands, indicating that it focuses on extracting features from these bands~\cite{lin2022investigating}; while RadImageNet pre-trained model responds more actively to low-frequency features. Similar trends are observed in the learning priorities after fine-tuning, as shown in Fig.~\ref{fig-03} (second row). Although the peaks eventually shift to higher frequencies, the overall PSDs still resemble their pre-trained counterparts. This is unsurprising, considering that kernels in early layers show minimal change during fine-tuning \cite{raghu2019transfusion}, thereby inheriting the predominant spectral response from pre-training.

\subsubsection{PSD is related to shortcut learning}
\begin{figure}[t]
    \centering
    \includegraphics[scale=1]{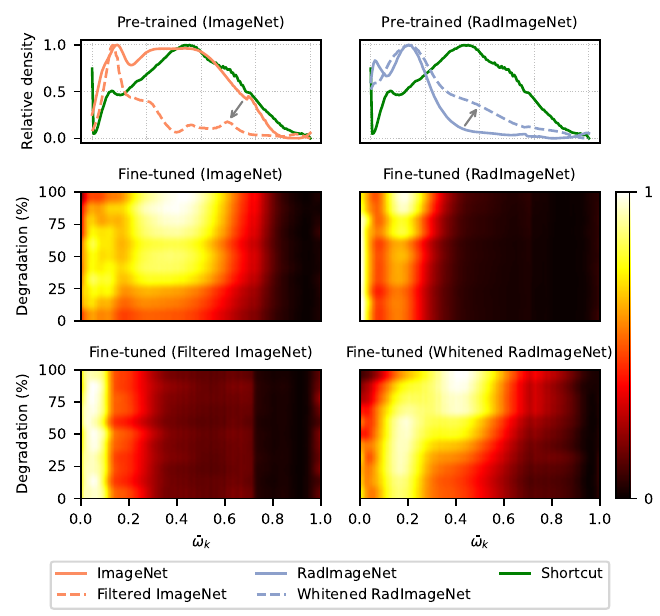}
    \caption{Normalized learning priorities of pre-trained and fine-tuned models. $\bar{\omega}_{k}$ is the normalized radical frequency with respect to the highest value (x-axis shared between rows). \textbf{Top}: normalized PSDs from Eq. 1. The arrows show how the pre-trained model PSDs change before and after source data editing. \textbf{Middle}: PSD as a heat map, after different degrees of degradation (amount of artifacts in the training set) for the original datasets. \textbf{Bottom}: Same as above but for the edited datasets.}
    \label{fig-03}
\end{figure}

We computed the average PSDs of artificially generated artifacts by extracting the residual between the original and modified images, plotted in Fig.\ref{fig-03} (green solid lines). We observe that the spectral distribution of the artifacts mainly falls in the mid-to-high frequencies. Interestingly, the learning priority of ImageNet pre-trained model shows a higher level of overlap with the PSD of the artifact, while the results in Fig.~\ref{fig-02} indicate that ImageNet is prone to shortcut learning. As gradients reflect how much the loss is affected by changes in the input, higher density indicates that kernels are more sensitive to corresponding frequency perturbations \cite{rahaman2019spectral}. Therefore, it is reasonable to believe that the learning priority of a pre-trained model and its robustness to frequency shortcuts are related: kernels pre-trained on ImageNet have stronger response to mid-to-high frequencies and thus can quickly detect shortcuts with similar spectral distributions.

\subsubsection{Source data affects robustness}
Previous experiments show that the frequency response of the early layers remains largely unchanged in transfer learning, thus it is possible to enhance or reduce shortcut learning by modifying the model's learning priority via source data editing. Specifically, we altered the model's response to mid-to-high frequencies during pre-training. We encouraged RadImageNet model to focus more on learning high frequencies by normalizing the spectrum of images in RadImageNet. Additionally, whitening was applied to ensure that the normalized images maintain the same mean and standard deviation as the originals:

\begin{equation}
    I_{n} = \mathcal{F}^{-1}\left ( \frac{\mathcal{F}\left ( I \right )}{\left \| \mathcal{F}\left ( I \right ) \right \|} \right ), \quad I_{w} = \left (I_{n} - \mu_{n}  \right )\frac{\sigma_{o}}{\sigma_{n}} + \mu_{o},
    \label{eq-2}
\end{equation}
where $I$, $I_{n}$, and $I_{w}$ represent the original, normalized, and whitened images, respectively. $\mu_{o}$, $\mu_{n}$ and $\sigma_{o}$, $\sigma_{n}$ are the mean and standard deviation of the original image and the normalized image, respectively. $\mathcal{F}^{-1}$ denotes the inverse Fourier transform.

\begin{figure}[t]
    \centering
    \includegraphics[scale=1]{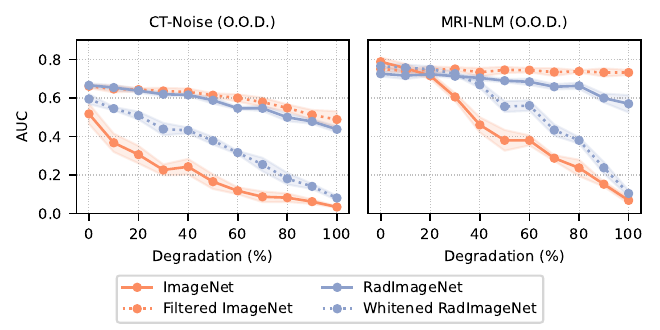}
    \caption{O.O.D. performance (mean and standard deviation of AUC across 5-folds) with (doted lines) and without (solid lines) source data editing.}
    \label{fig-04}
\end{figure}

On the contrary, we constrained ImageNet model to exclusively learn low-frequency patterns by eliminating high-frequency details from the ImageNet images. Due to the missing fine details between sub-classes, we merged similar classes based on hierarchy, reducing the number of classes to three: \textit{living thing}, \textit{artifact}, and \textit{miscellaneous}, to guarantee convergence. The performance of models pre-trained on modified datasets is illustrated in Fig.~\ref{fig-04}, with their learning priorities in Fig.~\ref{fig-03} (third row).

As expected, the model pre-trained on whitened RadImageNet no longer shows low learning priority in high frequencies and picks up the shortcut during fine-tuning. In contrast, the model pre-trained on filtered ImageNet has limited capability to learn high-frequency features, resulting in a learning priority similar to that of the model pre-trained on original RadImageNet and thereby achieving comparable or even improved robustness.

\section{Conclusions and Future Work}
\label{sec:concl}
In this paper, we discovered that a model's response to frequency shortcuts in transfer learning is influenced by the similarity between the spectral distribution of the shortcut and the learning priority of the pre-trained model. By modifying source data, we showed that it is possible to alter the fine-tuned model robustness against frequency shortcuts. Although frequency analysis is a promising technique for understanding model robustness in transfer learning, several questions remain. First, it is unclear how the statistics of the untouched source data may affect the model's learning priority during pre-training. Second, although we showed that fine-tuned model robustness can be altered, a fine-grained method to manipulate the model's PSD is preferred. Lastly, it would also be interesting to investigate other types of non-frequency confounders, such as patient gender, medical equipment, or markers, from the perspective of the frequency domain.

\section*{Acknowledgement}
This research was supported by National Institutes of Health under Grant NIH EY07977, and Novo Nordisk Foundation under Grant NNF21OC0068816.

%
%

\bibliographystyle{splncs04}
\bibliography{refs-yucl}

\end{document}